\PassOptionsToPackage{unicode}{hyperref}
\PassOptionsToPackage{hyphens}{url}
\documentclass[
]{article}
\usepackage{amsmath,amssymb}
\usepackage{lmodern}
\usepackage{iftex}
\usepackage[margin=1in]{geometry} 
\usepackage{graphicx}
\ifPDFTeX
  \usepackage[T1]{fontenc}
  \usepackage[utf8]{inputenc}
  \usepackage{textcomp} 
\else 
  \usepackage{unicode-math}
  \defaultfontfeatures{Scale=MatchLowercase}
  \defaultfontfeatures[\rmfamily]{Ligatures=TeX,Scale=1}
\fi
\IfFileExists{upquote.sty}{\usepackage{upquote}}{}
\IfFileExists{microtype.sty}{
  \usepackage[]{microtype}
  \UseMicrotypeSet[protrusion]{basicmath} 
}{}
\makeatletter
\@ifundefined{KOMAClassName}{
  \IfFileExists{parskip.sty}{%
    \usepackage{parskip}
  }{
    \setlength{\parindent}{0pt}
    \setlength{\parskip}{6pt plus 2pt minus 1pt}}
}{
  \KOMAoptions{parskip=half}}
\makeatother
\usepackage{xcolor}
\usepackage[normalem]{ulem}
\ifLuaTeX
  \usepackage{selnolig}  
\fi
\IfFileExists{bookmark.sty}{\usepackage{bookmark}}{\usepackage{hyperref}}
\IfFileExists{xurl.sty}{\usepackage{xurl}}{} 
\hypersetup{
  hidelinks,
  pdfcreator={LaTeX via pandoc}}

\PassOptionsToPackage{unicode}{hyperref}
\PassOptionsToPackage{hyphens}{url}
\ifPDFTeX
  \usepackage[T1]{fontenc}
  \usepackage{textcomp} 
\else 
  \usepackage{unicode-math}
  \defaultfontfeatures{Scale=MatchLowercase}
  \defaultfontfeatures[\rmfamily]{Ligatures=TeX,Scale=1}
\fi
\IfFileExists{upquote.sty}{\usepackage{upquote}}{}
\IfFileExists{microtype.sty}{
  \usepackage[]{microtype}
  \UseMicrotypeSet[protrusion]{basicmath} 
}{}
\makeatletter
\@ifundefined{KOMAClassName}{
  \IfFileExists{parskip.sty}{%
    \usepackage{parskip}
  }{
    \setlength{\parindent}{0pt}
    \setlength{\parskip}{6pt plus 2pt minus 1pt}}
}{
  \KOMAoptions{parskip=half}}
\makeatother
\usepackage{xcolor}
\usepackage{longtable,booktabs,array}
\usepackage{multirow}
\usepackage{calc} 
\usepackage{etoolbox}
\makeatletter
\patchcmd\longtable{\par}{\if@noskipsec\mbox{}\fi\par}{}{}
\makeatother
\IfFileExists{footnotehyper.sty}{\usepackage{footnotehyper}}{\usepackage{footnote}}
\makesavenoteenv{longtable}
\ifLuaTeX
  \usepackage{selnolig}  
\fi
\IfFileExists{bookmark.sty}{\usepackage{bookmark}}{\usepackage{hyperref}}
\IfFileExists{xurl.sty}{\usepackage{xurl}}{} 
\hypersetup{
  hidelinks,
  pdfcreator={LaTeX via pandoc}}

\usepackage{booktabs}
\usepackage{longtable}
\usepackage{array}
\usepackage{ragged2e} 
\usepackage{authblk} 
\title{Improving Radiology Report Conciseness and Structure via Local
Large Language Models}

\author[1]{Iryna Hartsock}
\author[2]{Cyrillo Araujo}
\author[3]{Les Folio}
\author[1]{Ghulam Rasool}

\affil[1]{Department of Machine Learning, H. Lee Moffitt Cancer Center \& Research Institute, Tampa, FL, USA}
\affil[2]{Deparmtent of Diagnostic Imaging \& Interventional Radiology, H. Lee Moffitt Cancer Center \& Research Institute, Tampa, FL, USA}
\affil[3]{Department of Radiology, James A. Haley VA Hospital, Tampa, FL, USA}

\date{}

\begin{document}

\maketitle

\textbf{Abstract}

Radiology reports are often lengthy and unstructured, posing challenges
for referring physicians to quickly identify critical imaging findings
while increasing risk of missed information. This retrospective study
aimed to enhance radiology reports by making them concise and
well-structured, with findings organized by relevant organs. To achieve
this, we utilized private large language models (LLMs) deployed locally
within our institution\textquotesingle s firewall, ensuring data
security and minimizing computational costs. Using a dataset of 814
radiology reports from seven board-certified body radiologists at
Moffitt Cancer Center, we tested five prompting strategies within
the LangChain framework. After evaluating several models, the Mixtral
LLM demonstrated superior adherence to formatting requirements compared
to alternatives like Llama. The optimal strategy involved condensing
reports first and then applying structured formatting based on specific
instructions, reducing verbosity while improving clarity. Across all
radiologists and reports, the Mixtral LLM reduced redundant word counts
by more than 53\%. These findings highlight the potential of locally
deployed, open-source LLMs to streamline radiology reporting. By
generating concise, well-structured reports, these models enhance
information retrieval and better meet the needs of referring physicians,
ultimately improving clinical workflows.

\textbf{Keywords:} radiology reports, large language models,
conciseness, structure

\textbf{\hfill\break
}

\textbf{Introduction}

A significant challenge in radiology reporting is that reports tend to
be overly verbose and poorly structured, making it~difficult~for
referring physicians to discern crucial findings and potentially
overlooking important information {[}1, 2{]}. Implementing structured
(templated) reporting methods provides a viable solution, enabling
physicians to access relevant information efficiently {[}3--5{]}.
Imaging findings can be structured in various ways, such as organizing
them from head-to-toe, prioritizing from the most critical to the least
important, and/or itemizing them by specific organs {[}6, 7{]}. Removing
redundancies, unnecessary words, and phrases from a radiology report
without compromising its meaning further enhances interpretation
efficiency {[}8, 9{]}. Ultimately, well-structured and concise radiology
reporting is not just about documentation; it is crucial for delivering
high-quality healthcare {[}2, 3, 10{]}.~

In recent years, numerous studies have explored using large language
models (LLMs) to improve the readability of radiology reports
{[}11--14{]}. LLMs are powerful artificial intelligence (AI) models
capable of analyzing and generating human-like natural language
{[}15--19{]}. For instance, Adams et al. employed Generative Pre-trained
Transformer 4 (GPT-4) to transform 170 free-text CT and MRI reports into
structured formats by selecting the best templates from a predefined
list, achieving successful conversions for all reports {[}11{]}. Jeblick
et al. used ChatGPT to simplify radiology reports suitable for a
child\textquotesingle s understanding {[}12{]}. While 15 radiologists
generally found the simplified reports to be factually accurate and not
harmful, some noted errors and overlooked details. Additionally, Mallio
et al. demonstrated the efficacy of ChatGPT-3.5 Turbo and GPT-4 models
in reducing the verbosity of radiology reports {[}13{]}. However, a key
limitation of these three studies is the use of application programming
interfaces (APIs) or LLMs provided as a service on the Internet, which
often involves sharing data with third parties or using
synthetically-generated radiology reports~{[}11--13{]}. Using external
LLMs raises concerns regarding data privacy, security, adherence~to
regulations such as the Health Insurance Portability and Accountability
Act (HIPAA), and growing usage costs {[}20, 21{]}.

In our IRB-approved study, we run state-of-the-art LLMs such as Mixtral
{[}15{]}, Mistral {[}16{]}, and Llama 3 {[}18{]} locally on a Windows
Desktop computer equipped with an ordinary graphical processing unit
(GPU), i.e., RTX 3060 with 12 Gig VRAM. This approach ensures that
patient data remains within our healthcare institution's secure
infrastructure, minimizing the risk of unauthorized access and
eliminating reliance on costly cloud-based LLM services. Our findings
demonstrate that locally deployed LLMs can significantly enhance
radiology reports by making them concise and well-structured, all while
maintaining data privacy and incurring minimal computational costs.

\hfill

\textbf{Materials and Methods}

\textbf{Radiology Reports}

In this retrospective quality improvement study, the IRB waived the
requirement for informed consent from the participating radiologists and
patients. We analyzed 814 radiology reports dictated by seven
board-certified body radiologists at our Cancer Center between 2023 and
2024. These reports were based on CT exams of the chest, abdomen, and
pelvis in patients with various cancers. Each report comprised two
standard sections: ``Findings,'' which detailed medical observations,
and ``Impressions,'' which summarized key findings. Report lengths
ranged from 182 to 981 words, with a mean of 372 words and a median of
344 words. Although most reports were organized by organs, variations in
radiologists' writing styles resulted in inconsistent formatting and
significant variations in the structure. To ensure patient
confidentiality, all reports were fully de-identified before analysis.

\textbf{Large Language Models (LLMs)}

The study employed the Mixtral 8x7B LLM, which utilizes a sparse
mixture-of-experts (SMoE) architecture with a total of 56 billion
parameters (i.e., eight experts, each with approximately 7B parameters)
{[}15{]}. For efficiency, we used pre-existing 6Q quantized weights of
Mixtral. In the Mixtral architecture, each layer incorporates eight
distinct feedforward expert blocks. During inference, based on the
specific characteristics of each token, two out of these eight experts
are selected to process the input, and thus, only 14B parameters are
active during inference. Mixtral benefits from a substantially large
context window of 32,768 tokens and was pre-trained on multilingual data
extracted from the open web {[}15{]}. Mixtral was selected over other
LLMs, including Mistral 7B {[}16{]} and Llama 3 8B {[}18{]}, due to its
superior adherence to a specific output formatting requirement compared
to other LLMs. All LLMs were run locally behind a secure firewall using
the Ollama framework to ensure data security and privacy {[}22{]}. The
temperature of the LLMs was set to zero to reduce randomness and ensure
more consistent outputs. The LLMs were downloaded from the Ollama store
and were not fine-tuned or updated in any manner.

\textbf{Prompt Engineering}

We utilized the LangChain library to prompt LLMs programmatically
{[}23{]}. Each radiology report underwent one or two calls to the LLM,
depending on the prompting strategy, to ensure both structure and
conciseness. The order of these steps varied across the strategies. The
prompts were developed iteratively in collaboration with board-certified
radiologists to align closely with clinical workflows. They included
instructions to remove redundant and non-meaningful words and phrases,
such as ``there is,'' ``of the,'' ``within the,'' ``visualized,''
``measures,'' ``approximately,'' ``the patient,'' and ``at this time,''
among others. A standardized template (depicted in Fig. 1) was used
across all strategies, designed with input from a radiologist to ensure
uniformity and clinical relevance.

The five prompting strategies described below, were refined to balance
conciseness and clarity while adhering to the standardized structure:

\begin{enumerate}
\def\labelenumi{\arabic{enumi}.}
\item
  \uline{Structure \textbf{{[}S{]}}}: This approach used a single prompt
  to organize the radiology report into the predefined template (see
  Supplementary Fig. 1).~
\item
  \uline{Structure \textgreater\textgreater{} Conciseness \textbf{{[}S
  \textgreater\textgreater{} C{]}}}: Reports were first structured using
  the predefined format via one prompt. A second prompt was then used to
  refine the structured output, enhancing conciseness while preserving
  the structure (Supplementary Fig. 2).~
\item
  \uline{Conciseness \textgreater\textgreater{} Structure \textbf{{[}C
  \textgreater\textgreater{} S{]}}}: In this strategy, the LLM first
  condensed the report and then organized the concise content into a
  predefined structure. This required two calls to the LLM
  (Supplementary Fig. 3).
\item
  \uline{Structure + Conciseness \textbf{{[}S + C{]}}}: This method
  combined structure and conciseness in a single prompt, instructing the
  LLM to address both aspects simultaneously (Supplementary Fig. 4).
\item
  \uline{Structure + Conciseness (F, I) {[}\textbf{S + C (F, I)}{]}}:
  This strategy involved separate prompts for the ``Findings'' and
  ``Impressions'' sections. Each prompt focused on structuring and
  condensing its respective section (Supplementary Fig. 5).
\end{enumerate}

Occasionally, LLMs failed to adhere to the required format, causing the
Python program to terminate unexpectedly. To mitigate this, we
incorporated an \textquotesingle OutputFixingParser,\textquotesingle{}
which allowed the LLM a second attempt to correct formatting errors. The
parser resubmitted the output alongside the original instructions,
ensuring better adherence to the desired structure. The source code is
available in our GitHub repository:
\url{https://github.com/lab-rasool/Radiology-reporting-with-LLMs}.

\textbf{Evaluation Using Word Count and Conciseness Score}

The word count of radiology reports may serve as a fundamental metric
for evaluating their verbosity and efficiency. A high word count may
indicate excessive use of redundant or unnecessary language, potentially
hindering readability and delaying the identification of key findings.
Conversely, a very low word count might suggest the omission of
essential clinical details, which could compromise diagnostic accuracy.
By examining word count, we gain insight into the overall length of
reports, allowing for comparisons between original and processed
versions.

To complement word count analysis, we calculate a \emph{conciseness
score}, which quantifies the percentage reduction in word count after
processing by LLMs. This score reflects the LLM's ability to streamline
reports while maintaining their informational integrity. The conciseness
score is computed using the following formula:

\[Conciseness\ score = \ \frac{Word\ count\ in\ \ LLM\ generated\ report}{Word\ count\ in\ \ original\ report}\  \times \ 100\%.\]

A conciseness score near 100\% indicates minimal modification to the
original report, with few or no words removed. A lower conciseness score
suggests a significant reduction in verbosity, implying that the
original report was less concise and required substantial editing by the
LLM. Conversely, a score above 100\% occurs when the LLM introduces
additional words, resulting in a lengthened report rather than a
condensed one.

Together, word count and conciseness score provide complementary
insights. Word count alone captures the absolute length of reports,
while the conciseness score highlights the efficiency of the LLM in
reducing unnecessary content. However, it is important to note that
neither metric fully addresses the clinical relevance of the content.
Essential details may increase word count but are crucial for ensuring
diagnostic utility. Additionally, variations in prompts or LLMs may
result in different word counts and conciseness scores for the same
report, reflecting differences in processing strategies and adherence to
formatting instructions. We acknowledge that conciseness and word count
may not be the only useful metrics; the comprehensibility of the
radiology report on the part of referring providers and/or patients is
also important.

\hfill
\newpage

\textbf{Results}

\textbf{LLMs Formatting and Processing Errors}

Two types of formatting errors were observed during the processing of
radiology reports by the LLMs: (1) \emph{Failure to structure reports} -
the LLM did not organize reports into the predefined format (Fig. 1)
after two attempts, and (2) \emph{Impression list generation errors} --
The LLM generated impressions as a sequence of individual letters
followed by new lines instead of complete impressions in bullet points
or paragraphs. This led to excessively long reports with a word count of
more than the original report. Reports with such errors were excluded
from further analysis. Formatting issues were often specific to
particular prompting methods, allowing selective exclusion of affected
cases.

As shown in Table 1, the {[}S + C (F, I){]} strategy demonstrated the
best performance, with only 23 reports (2.8\%) having formatting errors.
By using separate prompts for the ``Findings'' and ``Impressions''
sections, this strategy minimized errors. Conversely, the {[}S
\textgreater\textgreater{} C{]} strategy had the highest error rate,
with 181 reports (22.3\%) affected. This is likely due to the
compounding of errors from its chained prompts, where each prompt
enforces the LLM to follow formatting instructions. Other prompting
strategies encountered formatting errors in 58 to 88 reports (7-11\%).

Among the LLMs tested, Mixtral exhibited the lowest rate of formatting
errors. For example, in the {[}C \textgreater\textgreater{} S{]}
strategy, Mixtral produced errors in 88 reports (10.8\%), compared to
107 reports (13.1\%) for Mistral 7B and 726 reports (89.2\%) for Llama 3
8B. These results underscore Mixtral's robustness in adhering to
formatting instructions.

As summarized in Table 2, reports with conciseness scores exceeding
100\% (indicating longer LLM-generated reports) were relatively rare.
The {[}S + C (F, I){]} strategy had the highest occurrence of such
cases, with 36 reports (4.4\%). The {[}C \textgreater\textgreater{} S{]}
strategy had no reports exceeding 100\%, indicating strong adherence to
instructions. Reports with scores above 100\% often resulted from the
LLM over-listing findings, particularly in the ``Impressions'' section
(Fig. 5B--D).

\textbf{Performance of the {[}C \textgreater\textgreater{} S{]}
Prompting Strategy}

To evaluate the effectiveness of the Mixtral LLM in improving radiology
reports, we used the {[}C \textgreater\textgreater{} S{]} prompting
strategy as a representative example, as it well adhered to formatting
instructions. This strategy focuses on condensing reports first,
followed by organizing the concise content into a predefined structure.
Fig. 2 highlights the performance of the {[}C \textgreater\textgreater{}
S{]} strategy by comparing word counts of the original and LLM-processed
reports across multiple radiologists. The first subplot presents the
overall average word count, with error bars representing one standard
deviation. Subsequent subplots showcase word count comparisons for 15
randomly selected reports per radiologist, emphasizing the variability
in length reduction achieved. The results show that the {[}C
\textgreater\textgreater{} S{]} strategy consistently reduced word
counts across all radiologists. Before LLM processing, the average word
count ranged from 225.92 to 509.91, reflecting significant variability
in initial verbosity across radiologists. After LLM processing, the
average word count was significantly reduced, ranging from 136.75 to
178.45, demonstrating the strategy\textquotesingle s ability to condense
reports effectively. Standard deviations also decreased after
processing, indicating improved consistency in report length. For
instance, for radiologist 1, the standard deviation reduced from 161.08
(before) to 62.85 (after) and for radiologist 3, the standard deviation
reduced from 110.46 (before) to 53.87 (after).

These reductions highlight the robustness of the {[}C
\textgreater\textgreater{} S{]} strategy in standardizing reports and
managing diverse writing styles. Although minor variations in word count
reductions were observed among radiologists, these differences align
with their original reporting styles and verbosity. However, it should
be noted that the reduction in word count does not necessarily imply
retention of critical clinical information.

\textbf{Comparison of Prompting Strategies for Enhancing Report
Conciseness}

We evaluated the performance of five prompting strategies, {[}S{]}, {[}S
\textgreater\textgreater{} C{]}, {[}C \textgreater\textgreater{} S{]},
{[}S + C{]}, and {[}S + C (F, I){]}, to assess their effectiveness in
improving radiology report conciseness. Each strategy was designed to
balance reducing verbosity with adhering to a standardized structure.
Outputs with conciseness scores over 100\% were excluded from this
analysis. Fig. 3 illustrates the distribution of conciseness scores (\%)
for LLM-processed reports across radiologists for the five strategies.
Each box plot represents the median, interquartile range, and
variability of scores for a specific strategy grouped by radiologists.
This visualization highlights how both the prompting strategy and
individual reporting styles influence the LLM\textquotesingle s
performance. The figure reveals consistent trends in concise report
writing among radiologists, with radiologist 1 producing the least
concise reports and radiologist 7 the most concise, regardless of the
prompting method.

Similarly, Fig. 4 provides box plots for each radiologist, showing the
distribution of conciseness scores (\%) across the strategies. The
effect of prompt strategies on the conciseness scores was
radiologist-dependent, as reflected in their median values. Across the
seven radiologists, the choice of prompt strategy had a notable impact
on conciseness scores. For all radiologists, the {[}S
\textgreater\textgreater{} C{]} strategy resulted in the lowest median
conciseness scores. The {[}S \textgreater\textgreater{} C{]} strategy
also generally exhibited lower variability in scores, as measured by the
interquartile range (IQR), compared to the other strategies, indicating
more stable performance across radiologists.

The non-parametric Kruskal-Wallis test for each radiologist yielded
p-values \textless{} 0.001, indicating significant differences in
conciseness scores across strategies. To further analyze these
differences, we conducted Dunn's test with a Bonferroni correction of 70
for pairwise comparisons, highlighting results below the 0.05
significance threshold in Fig. 4. It confirmed that the {[}S
\textgreater\textgreater{} C{]} approach demonstrated statistically
significantly lower conciseness scores compared to the other strategies,
while the remaining strategies exhibited no significant differences in
overall performance. However, while the lower conciseness scores of the
{[}S \textgreater\textgreater{} C{]} approach are desirable, they come
with a higher likelihood of omitting critical clinical information.

\textbf{Streamlining an Unstructured Radiology Report: A Representative
Case Study}

To evaluate the Mixtral LLM's ability to structure and condense
unstructured radiology reports, we applied it to an example case (Fig.
5A). Using five distinct prompting strategies, the LLM generated
structured and concise reports (Fig. 5B--F), reducing the word count
from 388 to between 190 and 242 words depending on the strategy. Among
these, the {[}S{]} prompting approach, which focused solely on
formatting, produced the least condensed report. In contrast, the {[}C
\textgreater\textgreater{} S{]} and {[}S + C (F, I){]} approaches
generated the most concise outputs.

Significant variations were observed in how Mixtral handled the
``Impressions'' section. The {[}C \textgreater\textgreater{} S{]} and
{[}S + C (F, I){]} strategies adhered to instructions by directly
condensing content from the ``Impressions'' section. However, other
strategies relied more heavily on the ``Findings'' section to generate
``Impressions,'' deviating from the intended focus.

While Mixtral generally processed the ``Findings'' section consistently,
discrepancies were noted. For example, when instructed to classify
findings as ``Unremarkable'' for normal and ``None'' for absent, Mixtral
occasionally misclassified findings. As shown in Fig. 5B, E--F, the
model incorrectly reported no findings for the hepatobiliary system and
adrenals, likely due to oversight or misclassification. Additionally,
Mixtral sometimes attributed the same clinical findings to multiple
organs. For instance, ``general osteopenia with moderate degenerative
changes in the thoracic and lumbar spine'' was incorrectly listed under
abdominal, pelvic, and soft tissue findings (Fig. 3E). Other
misplacements included findings relevant to the mediastinum being
categorized as general chest findings.

\hfill

\textbf{Discussion}

This study demonstrates the potential of locally deployed LLMs to
enhance the conciseness and structure of radiology reports. Using a
standard Windows desktop with a GeForce RTX 3060 GPU (12 GB VRAM), we
evaluated state-of-the-art open-source LLMs, including Mixtral, Mistral,
and Llama. Among these, the Mixtral LLM emerged as the most reliable due
to its lower rate of formatting errors and superior adherence to prompt
instructions. To our knowledge, this is one of the first studies
exploring the use of LLMs in private, resource-constrained environments
for radiology reporting, emphasizing their feasibility for improving
clinical workflows while ensuring data security.

Our definition of ``structured reports'' refers to reports organized
into templated formats by organ systems, differing from the ACR RADS
definition that includes detailed management recommendations and
category-specific modules. The Mixtral LLM was evaluated across five
prompting strategies. The {[}S \textgreater\textgreater C{]} approach
produced the most concise report but resulted in the highest number of
formatting errors. While the {[}C \textgreater\textgreater{} S{]}
strategy generally produced less concise reports than the {[}S
\textgreater\textgreater{} C{]} approach, it proved to be more effective
overall due to its better adherence to formatting instructions and zero
\textquotesingle Impression\textquotesingle{} generation errors. The
{[}C \textgreater\textgreater{} S{]} method first condensed the reports,
reducing verbosity, and then applied a structured format, minimizing the
risk of omitting critical clinical details. The {[}S + C (F, I){]}
approach, which focused on separately structuring the ``Findings'' and
``Impressions'' sections, also performed well in reducing formatting
errors. These findings highlight that tailoring prompt strategies can
significantly improve the accuracy and reliability of LLM-generated
outputs.

We used two primary metrics to evaluate LLM performance: (1) the word
count, which assessed verbosity by comparing the report lengths before
and after processing, and (2) the conciseness score, which quantified
the percentage reduction in word count while preserving essential
information. These metrics provided valuable insights into the
efficiency of LLMs in reducing redundancy and structuring information.
However, they also underscored the importance of balancing brevity with
clinical relevance. Certain phrases or details, while increasing word
count, are essential for diagnostic clarity. Overemphasis on conciseness
might compromise the report's utility. Therefore, these metrics should
be supplemented with qualitative assessments of clinical accuracy.

This study builds upon prior research that has explored the use of LLMs
for radiology reporting. For instance, Adams et al. demonstrated the
feasibility of using GPT-4 to transform free-text reports into
structured formats, while Jeblick et al. utilized ChatGPT to simplify
reports for lay audiences {[}11, 12{]}. Unlike these studies, which
relied on API-based LLMs, our approach emphasized local deployment,
ensuring patient data security and cost-effectiveness. By using real
clinical data instead of synthetic datasets, our study addresses a
critical gap in the practical application of LLMs in healthcare
settings.

The findings have significant implications for radiology workflows.
Locally deployed LLMs, such as Mixtral, offer a cost-effective and
secure alternative to cloud-based models, addressing data privacy
concerns and compliance with HIPAA. By producing concise,
well-structured reports, these models can improve communication between
radiologists and referring physicians, enabling faster identification of
critical findings and enhancing patient care. Additionally, the
conciseness and structure provided by these models can serve as training
tools for radiology residents, fostering clarity and precision in
reporting.

While the results are promising, several limitations warrant
consideration. First, despite Mixtral\textquotesingle s robust
performance, formatting errors were observed in 23 to 181 reports across
prompting strategies. These errors were more frequent in multi-step
strategies like {[}S \textgreater\textgreater{} C{]}, which compounded
minor deviations across prompts. The LLM occasionally marked findings
for specific organs, such as the hepatobiliary system and adrenals, as
``None,'' particularly in unremarkable cases. These issues, though less
common with {[}C \textgreater\textgreater{} S{]}, highlight areas for
improvement in classification accuracy. In 61 instances, LLM-processed
reports exceeded the original length, contrary to the goal of
conciseness. These cases often involved overlisting findings,
particularly in the ``Impressions'' section. The study was limited to
radiology reports for CT scans of the chest, abdomen, and pelvis, and
the findings may not generalize to other medical domains or report
types. While prompt engineering was sufficient for this study, the lack
of fine-tuning limited the model\textquotesingle s ability to adapt to
specific clinical nuances.

To address the limitations and expand upon these findings, future
efforts could focus on fine-tuning LLMs using domain-specific datasets
to enhance their ability to handle complex clinical findings and reduce
misclassifications. Expanding the dataset to include reports from other
imaging modalities, such as MRI and ultrasound, as well as reports from
other clinical domains, would further improve generalizability.
Iterative feedback from radiologists could be leveraged to refine
prompts and outputs, thereby enhancing accuracy and usability.
Additionally, deploying LLMs in larger healthcare systems and diverse
workflows would provide valuable insights into their scalability and
broader applicability. Finally, a direct comparison between locally
deployed LLMs and API-based alternatives could help elucidate trade-offs
in performance, cost, and privacy, guiding informed decisions for
clinical adoption.

\hfill
\newpage

\textbf{Conclusion}

This study demonstrates the potential of locally deployed LLMs,
particularly the Mixtral model, to streamline radiology reporting by
generating concise and well-structured outputs. Among the prompting
strategies evaluated, the {[}C \textgreater\textgreater{} S{]} approach
emerged as the most effective, achieving significant reductions in
verbosity while maintaining clinical accuracy and structural adherence.
These findings highlight the feasibility of using LLMs in
resource-constrained and privacy-sensitive environments, providing a
secure, cost-effective alternative to cloud-based models. The
application of LLMs to radiology reporting addresses key challenges in
clinical workflows, including inconsistent formatting and overly verbose
reports, which can hinder the quick identification of critical findings
by referring physicians. By improving clarity and consistency, these
models have the potential to enhance communication, expedite patient
care, and support training for radiology residents. Despite its
strengths, this study also identified limitations, such as occasional
formatting errors, misclassifications, and the need for further
generalizability across other clinical domains.

\begin{figure}
    \centering
    \includegraphics[width=0.9\linewidth]{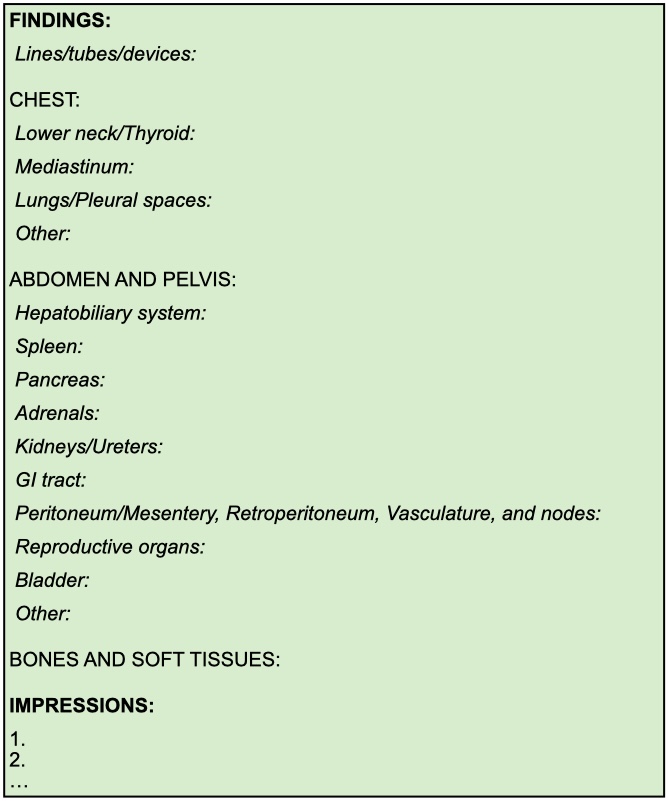}
    \caption{The structure of a radiology report, with organs and sub-organs ordered from head to toe, makes it intuitive and predictable for readers to locate the necessary information quickly. All five prompts proposed in this study use this structure as a template for the LLMs to follow when processing radiology reports. Thus, regardless of which radiologist writes the initial report, the LLM should generate a radiology report that adheres to this structured format.}
\end{figure}

\begin{figure}
    \centering
    \includegraphics[width=0.88\linewidth]{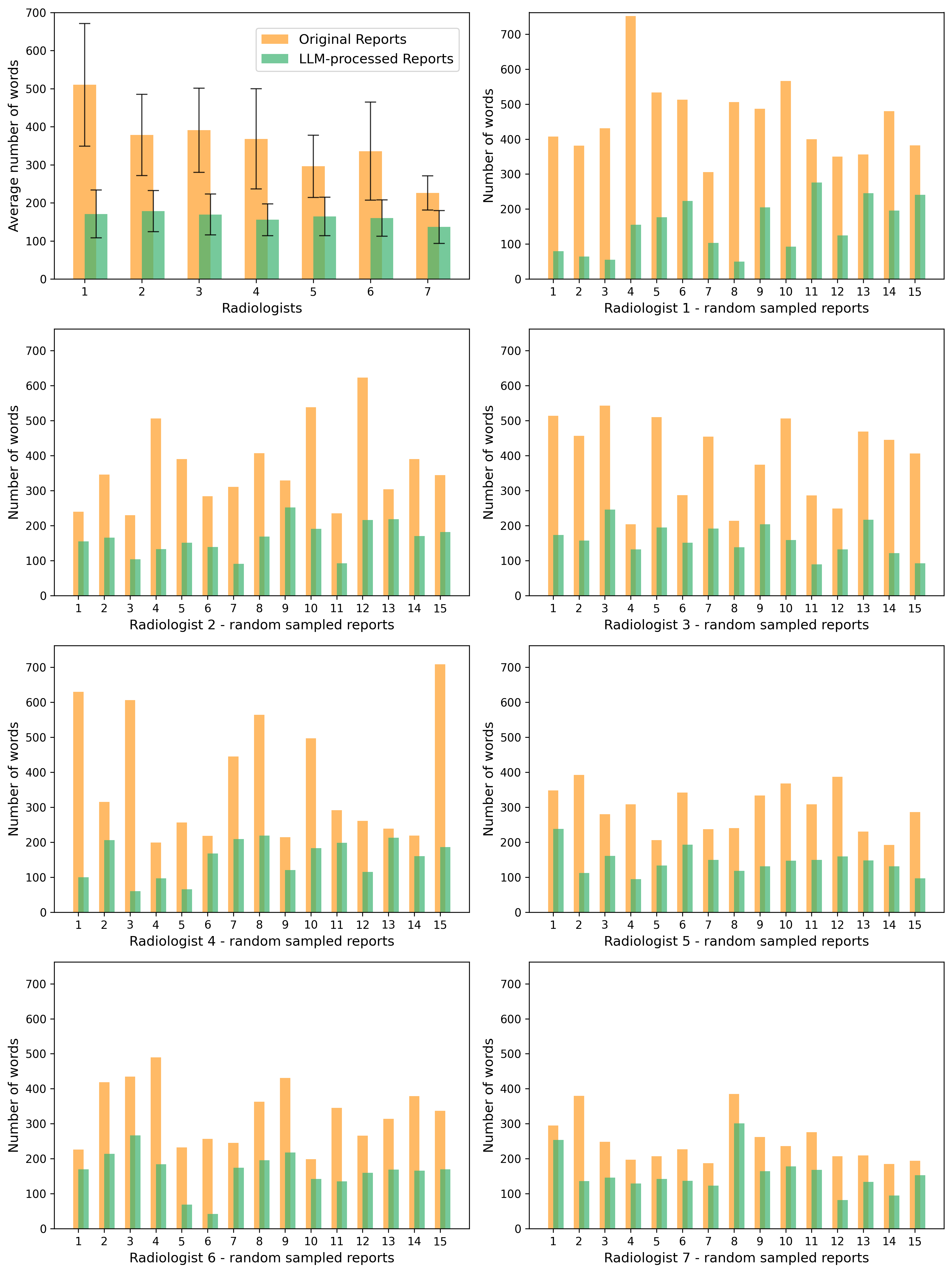}
    \caption{Comparison of word counts between original radiology reports and LLM-processed reports using the Conciseness >> Structure prompt strategy. The first subplot (top-left) displays the average word counts for original reports (orange) and LLM-processed reports (green) across all radiologists, with error bars representing one standard deviation. Each subsequent subplot represents a specific radiologist, showing the word counts for 15 randomly sampled reports, highlighting the variability and reduction in report length achieved by the LLM-processed reports. Radiologists are denoted by their respective indices, and the x-axis indicates the sampled reports.}
\end{figure}

\begin{figure}
    \centering
    \includegraphics[width=0.9\linewidth]{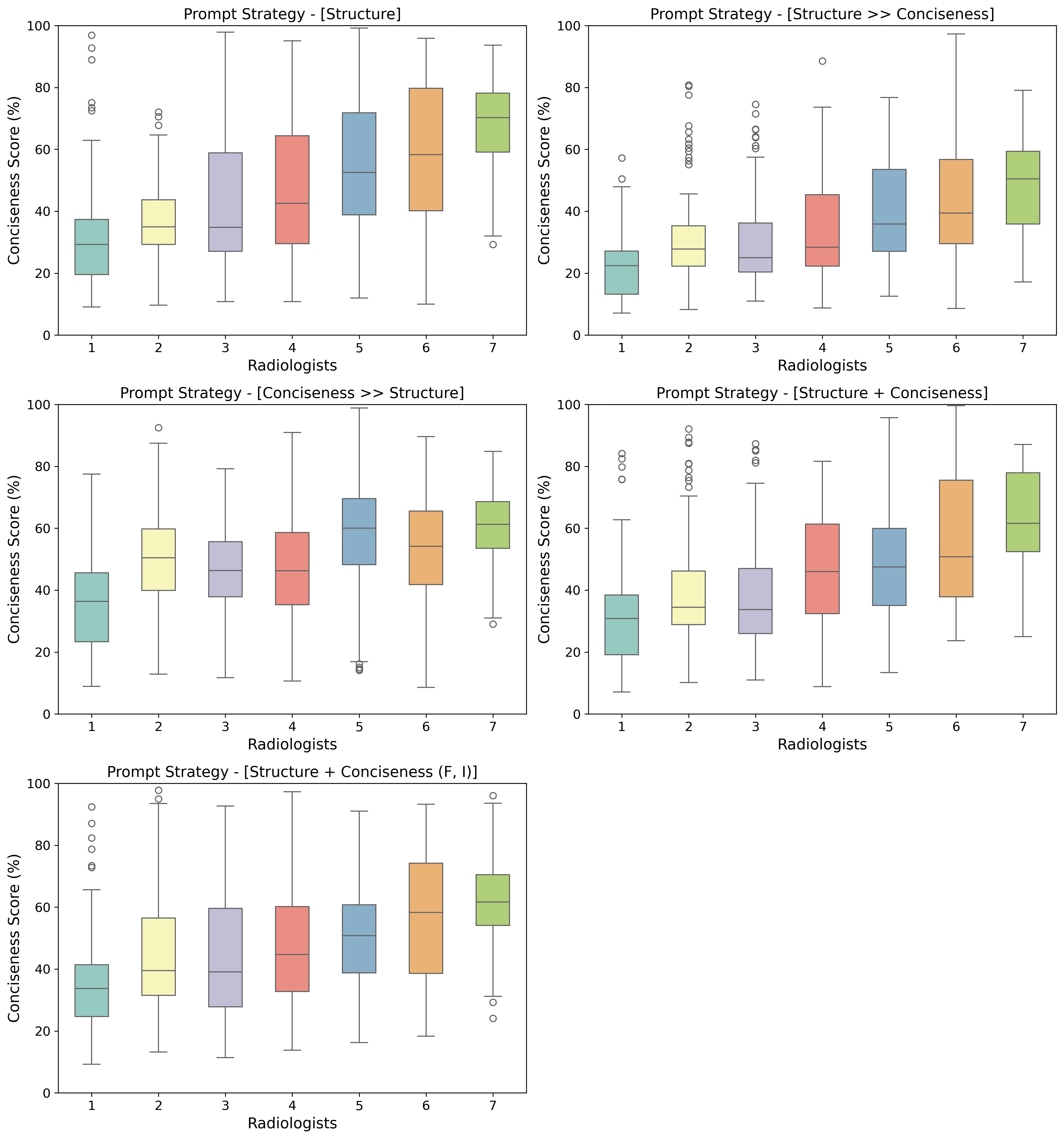}
    \caption{Box plots showing the conciseness scores (\%) of LLM-processed radiology reports for five prompting strategies: [Structure], [Structure >> Conciseness], [Conciseness >> Structure], [Structure + Conciseness], and [Structure + Conciseness (F, I)]. Each subplot represents the performance of a specific prompting strategy across reports generated by seven radiologists. Variability in median scores and interquartile ranges highlights the influence of prompting strategy and radiologist-specific factors on LLM performance.}
\end{figure}

\begin{figure}
    \centering
    \includegraphics[width=0.9\linewidth]{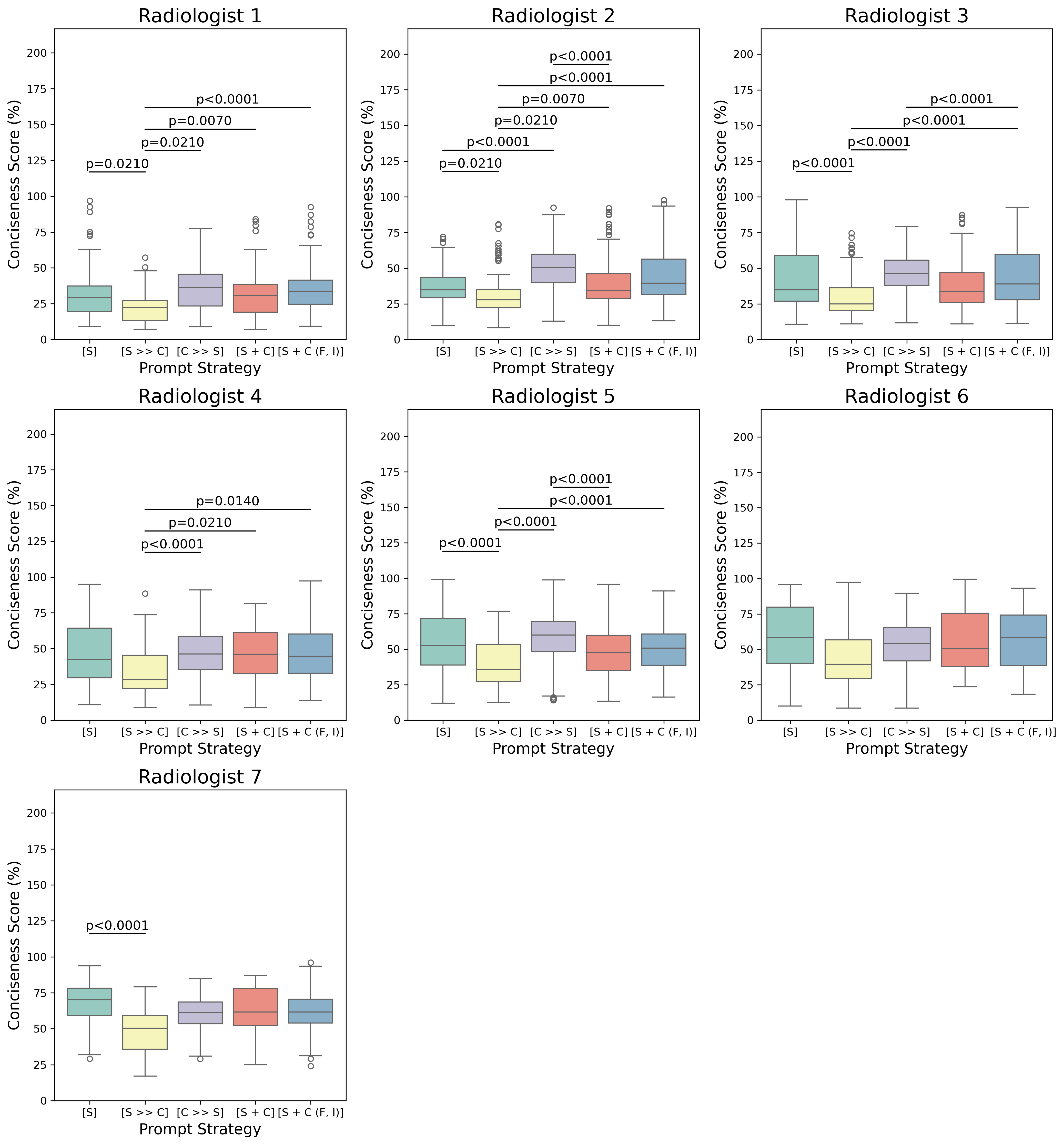}
    \caption{Fig. 4. Box plots comparing the conciseness scores (\%) of LLM-processed radiology reports for five prompting strategies across seven individual radiologists. Each subplot corresponds to a specific radiologist, with results displayed for the five strategies: [Structure], [Structure >> Conciseness], [Conciseness >> Structure], [Structure + Conciseness], and [Structure + Conciseness (F, I)]. The results highlight the variability in LLM performance for different strategies and radiologists, demonstrating the interplay between strategy effectiveness and radiologist-specific writing styles. Pairs of prompting strategies that are statistically significantly different are marked with horizontal bars, and their Bonferroni-corrected p-values are shown.}
\end{figure}

\begin{figure}
    \centering
    \includegraphics[width=0.9\linewidth]{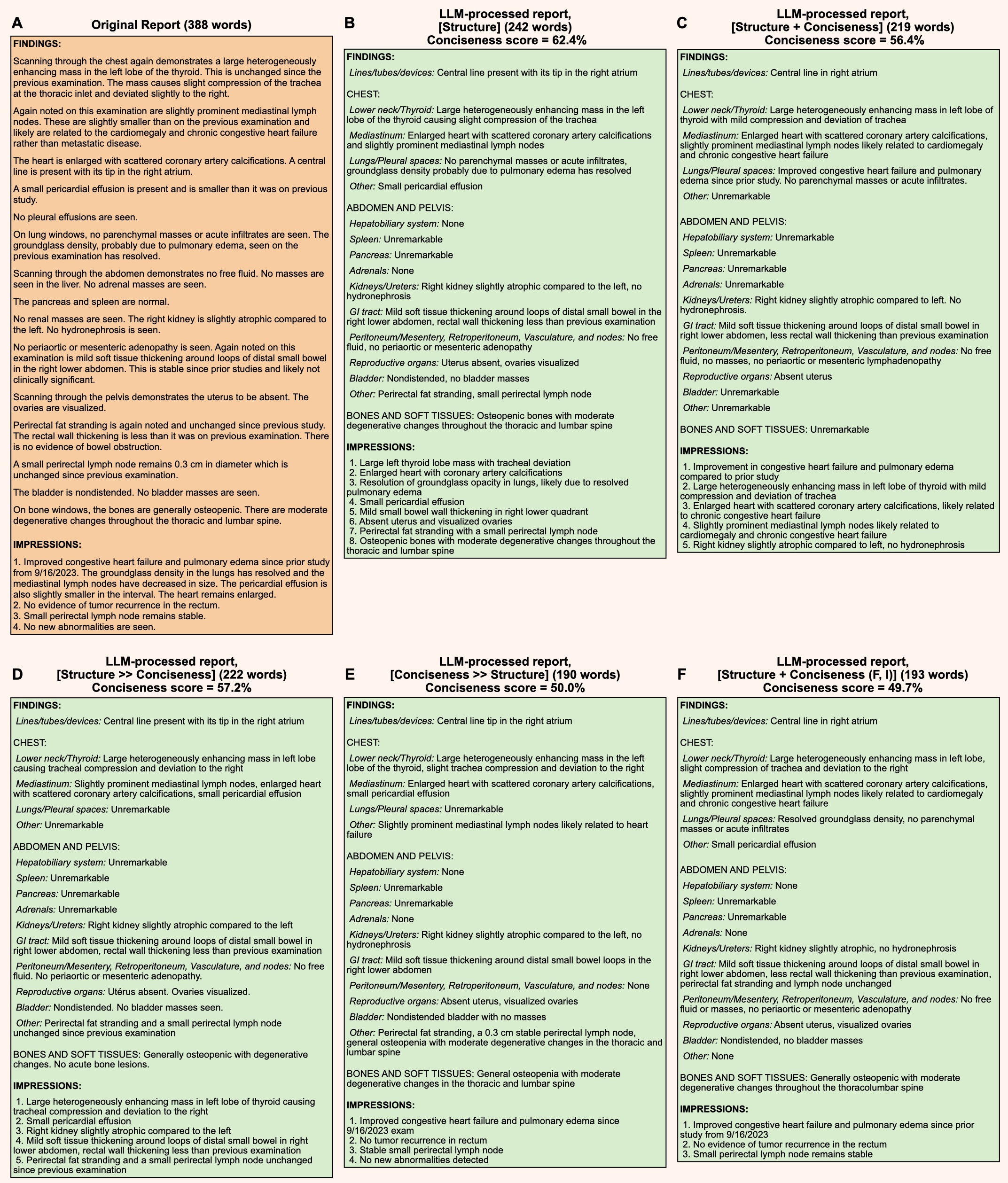}
    \caption{The performance of the Mixtral LLM on making the unstructured radiology report concise and well-structured under various prompting strategies. A Unstructured radiology report alongside its corresponding LLM-processed versions using the following prompting approaches: B [Structure], C [Structure + Conciseness], D [Structure >> Conciseness], E [Conciseness >> Structure], and F [Structure + Conciseness (F, I)]. The word count for each report is provided in parentheses. The conciseness scores of all LLM-processed reports are also indicated, with reports E and F having the lowest conciseness scores, making them the most concise. }
\end{figure}

\newpage 

\begin{longtable}{@{}%
  >{\raggedright\arraybackslash}p{0.12\linewidth}%
  >{\centering\arraybackslash}p{0.08\linewidth}%
  >{\centering\arraybackslash}p{0.08\linewidth}%
  >{\centering\arraybackslash}p{0.08\linewidth}%
  >{\centering\arraybackslash}p{0.08\linewidth}%
  >{\centering\arraybackslash}p{0.08\linewidth}%
  >{\centering\arraybackslash}p{0.08\linewidth}@{}}
\caption{The total number of radiology reports processed without formatting errors by the Mixtral LLM across five prompting approaches and seven body radiologists. The last row displays the total number of radiology reports without formatting errors for each approach and its percentage of the overall reports. The {[}Structure + Conciseness (F,I){]} prompting strategy had the fewest formatting errors.} \\
\toprule
\textbf{Radiologist} & \textbf{\# Reports} & \textbf{[S]} & \textbf{[S $\gg$ C]} & \textbf{[C $\gg$ S]} & \textbf{[S + C]} & \textbf{[S + C (F,I)]} \\
\midrule
\endfirsthead

\toprule
\textbf{Radiologist} & \textbf{\#Reports} & \textbf{[S]} & \textbf{[S $\gg$ C]} & \textbf{[C $\gg$ S]} & \textbf{[S + C]} & \textbf{[S + C (F,I)]} \\
\midrule
\endhead

Radiologist 1 & 111 & 103 & 85 & 82 & 99 & 104 \\
Radiologist 2 & 151 & 148 & 122 & 138 & 140 & 148 \\
Radiologist 3 & 126 & 118 & 102 & 119 & 114 & 123 \\
Radiologist 4 & 96  & 83  & 73  & 88  & 92  & 93  \\
Radiologist 5 & 192 & 183 & 143 & 191 & 175 & 186 \\
Radiologist 6 & 80  & 71  & 59  & 55  & 67  & 80  \\
Radiologist 7 & 58  & 50  & 49  & 53  & 52  & 57  \\
\textbf{Total} & \textbf{814} & 
\textbf{756 (92.9\%)} & 
\textbf{633 (77.8\%)} & 
\textbf{726 (89.2\%)} & 
\textbf{739 (90.8\%)} & 
\textbf{791 (97.2\%)} \\
\bottomrule
\end{longtable}

\begin{longtable}{@{}%
  >{\raggedright\arraybackslash}p{0.12\linewidth}%
  >{\centering\arraybackslash}p{0.08\linewidth}%
  >{\centering\arraybackslash}p{0.08\linewidth}%
  >{\centering\arraybackslash}p{0.08\linewidth}%
  >{\centering\arraybackslash}p{0.08\linewidth}%
  >{\centering\arraybackslash}p{0.08\linewidth}%
  >{\centering\arraybackslash}p{0.08\linewidth}@{}}
\caption{The total number of radiology reports processed by the Mixtral LLM with a conciseness score greater than 100\%, indicating that the LLM-processed report is longer than the original report, is summarized across five prompting approaches and seven body radiologists. The last row displays the total number of radiology reports with a conciseness score greater than 100\% for each approach and its percentage of the overall reports. Only a small number of radiology reports have a conciseness score over 100\%, with no such cases in the [Conciseness >> Structure] prompting strategy. } \\
\toprule
\textbf{Radiologist} & \textbf{\# Reports} & \textbf{[S]} & \textbf{[S $\gg$ C]} & \textbf{[C $\gg$ S]} & \textbf{[S + C]} & \textbf{[S + C (F,I)]} \\
\midrule
\endfirsthead

\toprule
\textbf{Radiologist} & \textbf{\#Reports} & \textbf{[S]} & \textbf{[S $\gg$ C]} & \textbf{[C $\gg$ S]} & \textbf{[S + C]} & \textbf{[S + C (F,I)]} \\
\midrule
\endhead

Radiologist 1 & 111 & 1 & 0 & 0 & 0 & 2 \\
Radiologist 2 & 151 & 0 & 0 & 0 & 0 & 1 \\
Radiologist 3 & 126 & 3 & 0 & 0 & 0 & 10 \\
Radiologist 4 & 96  & 1  & 0 & 0  & 0  & 11  \\
Radiologist 5 & 192 & 7 & 0 & 0 & 2 & 5 \\
Radiologist 6 & 80  & 3  & 2  & 0  & 4  & 6  \\
Radiologist 7 & 58  & 1  & 0  & 0  & 1  & 1  \\
\textbf{Total} & \textbf{814} & 
\textbf{16 (2.0\%)} & 
\textbf{2 (0.2\%)} & 
\textbf{0 (0.0\%)} & 
\textbf{7 (0.9\%)} & 
\textbf{36 (4.4\%)} \\
\bottomrule
\end{longtable}

\textbf{\hfill\break
}

\newpage 

\textbf{References}

1. Wallis A, McCoubrie P (2011) The radiology report --- Are we getting
the message across? Clin Radiol 66:1015--1022.
https://doi.org/10.1016/j.crad.2011.05.013

2. Franconeri A, Fang J, Carney B, et al (2018) Structured vs narrative
reporting of pelvic MRI for fibroids: clarity and impact on treatment
planning. Eur Radiol 28:3009--3017.
https://doi.org/10.1007/s00330-017-5161-9

3. Jorg T, Heckmann JC, Mildenberger P, et al (2021) Structured
reporting of CT scans of patients with trauma leads to faster, more
detailed diagnoses: An experimental study. Eur J Radiol 144:109954.
https://doi.org/10.1016/j.ejrad.2021.109954

4. Kim SH, Sobez LM, Spiro JE, et al (2020) Structured reporting has the
potential to reduce reporting times of dual-energy x-ray absorptiometry
exams. BMC Musculoskelet Disord 21:248.
https://doi.org/10.1186/s12891-020-03200-w

5. Schwartz LH, Panicek DM, Berk AR, et al (2011) Improving
Communication of Diagnostic Radiology Findings through Structured
Reporting. Radiology 260:174--181.
https://doi.org/10.1148/radiol.11101913

6. Nobel JM, Kok EM, Robben SGF (2020) Redefining the structure of
structured reporting in radiology. Insights Imaging 11:10.
https://doi.org/10.1186/s13244-019-0831-6

7. Sistrom CL, Langlotz CP (2005) A framework for improving radiology
reporting. Journal of the American College of Radiology 2:159--167.
https://doi.org/10.1016/j.jacr.2004.06.015

8. Hartung MP, Bickle IC, Gaillard F, Kanne JP (2020) How to Create a
Great Radiology Report. RadioGraphics 40:1658--1670.
https://doi.org/10.1148/rg.2020200020

9. Webster Riggs (2015) Making a case for concise narrative radiology
reports. Appl Radiol 44:4--6

10. Schoeppe F, Sommer WH, Nörenberg D, et al (2018) Structured
reporting adds clinical value in primary CT staging of diffuse large
B-cell lymphoma. Eur Radiol 28:3702--3709.
https://doi.org/10.1007/s00330-018-5340-3

11. Adams LC, Truhn D, Busch F, et al (2023) Leveraging GPT-4 for Post
Hoc Transformation of Free-text Radiology Reports into Structured
Reporting: A Multilingual Feasibility Study. Radiology 307:.
https://doi.org/10.1148/radiol.230725

12. Jeblick K, Schachtner B, Dexl J, et al (2023) ChatGPT makes medicine
easy to swallow: an exploratory case study on simplified radiology
reports. Eur Radiol 34:2817--2825.
https://doi.org/10.1007/s00330-023-10213-1

13. Mallio CA, Bernetti C, Sertorio AC, Zobel BB (2024) ChatGPT in
radiology structured reporting: analysis of ChatGPT-3.5 Turbo and GPT-4
in reducing word count and recalling findings. Quant Imaging Med Surg
14:2096--2102. https://doi.org/10.21037/qims-23-1300

14. Butler JJ, Puleo J, Harrington MC, et al (2024) From technical to
understandable: Artificial Intelligence Large Language Models improve
the readability of knee radiology reports. Knee Surgery, Sports
Traumatology, Arthroscopy 32:1077--1086.
https://doi.org/10.1002/ksa.12133

15. Jiang AQ, Sablayrolles A, Roux A, et al (2024) Mixtral of Experts

16. Jiang AQ, Sablayrolles A, Mensch A, et al (2023) Mistral 7B

17. Touvron H, Martin L, Stone K, et al (2023) Llama 2: Open Foundation
and Fine-Tuned Chat Models

18. Dubey A, Jauhri A, Pandey A, et al (2024) The Llama 3 Herd of Models

19. OpenAI, Achiam J, Adler S, et al (2023) GPT-4 Technical Report

20. Rosenbloom ST, Smith JRL, Bowen R, et al (2019) Updating HIPAA for
the electronic medical record era. Journal of the American Medical
Informatics Association 26:1115--1119.
https://doi.org/10.1093/jamia/ocz090

21. Hjerppe K, Ruohonen J, Leppanen V (2019) The General Data Protection
Regulation: Requirements, Architectures, and Constraints. In: 2019 IEEE
27th International Requirements Engineering Conference (RE). IEEE, pp
265--275

22. Ollama. https://ollama.com. Accessed 13 Oct 2024

23. LangChain. https://www.langchain.com. Accessed 13 Oct 2024

~

\newpage
\textbf{Supplementary Material}
\setcounter{figure}{0}
\begin{figure}[h]
    \centering
    \includegraphics[width=0.85\linewidth]{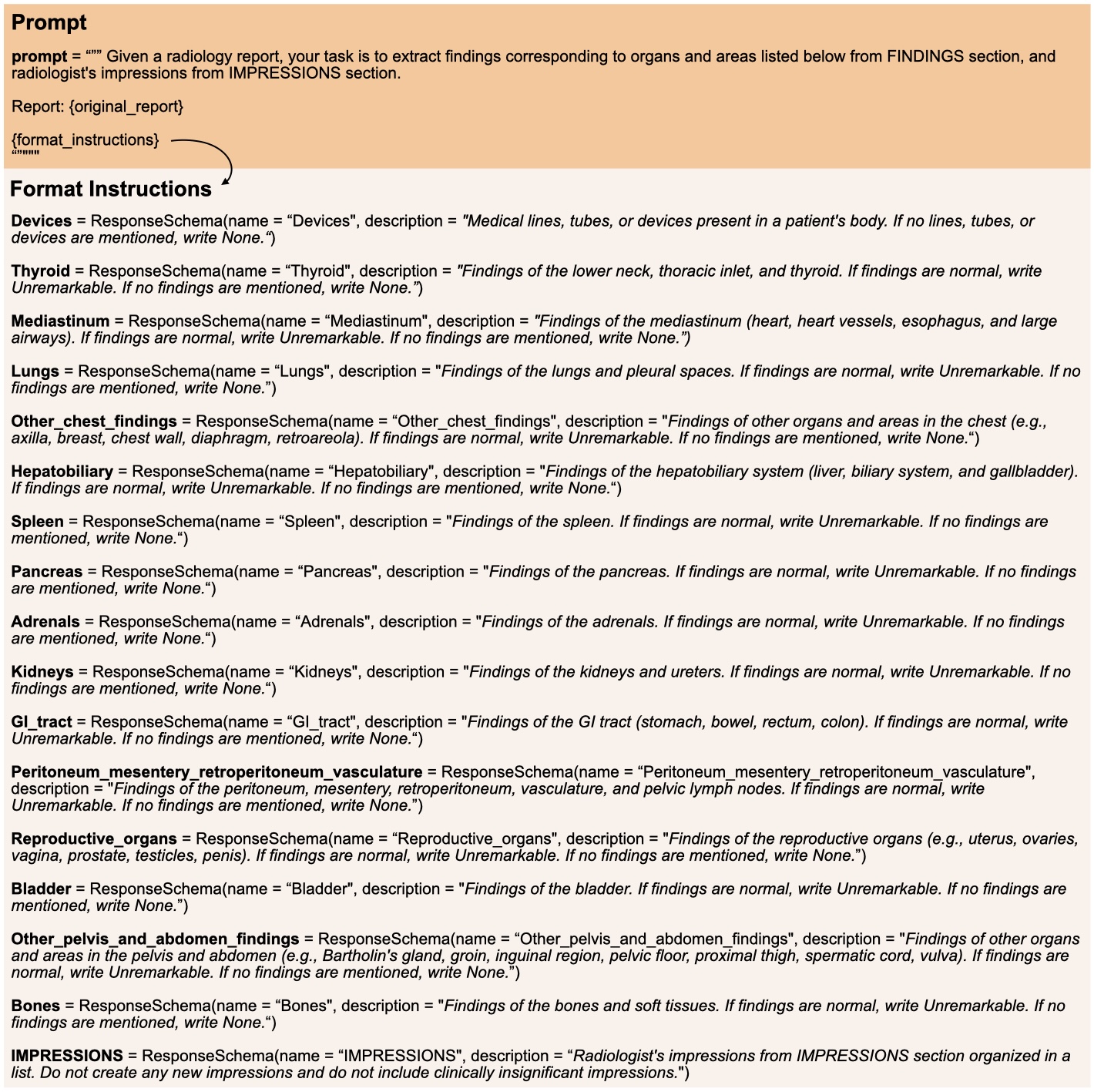}
    \caption{The prompt and formatting instructions used in the [Structure] prompting approach. The LLM is only prompted to structure the given radiology report according to specific formatting guidelines, as shown in Fig. 1.}
\end{figure}

\begin{figure}
    \centering
    \includegraphics[width=0.9\linewidth]{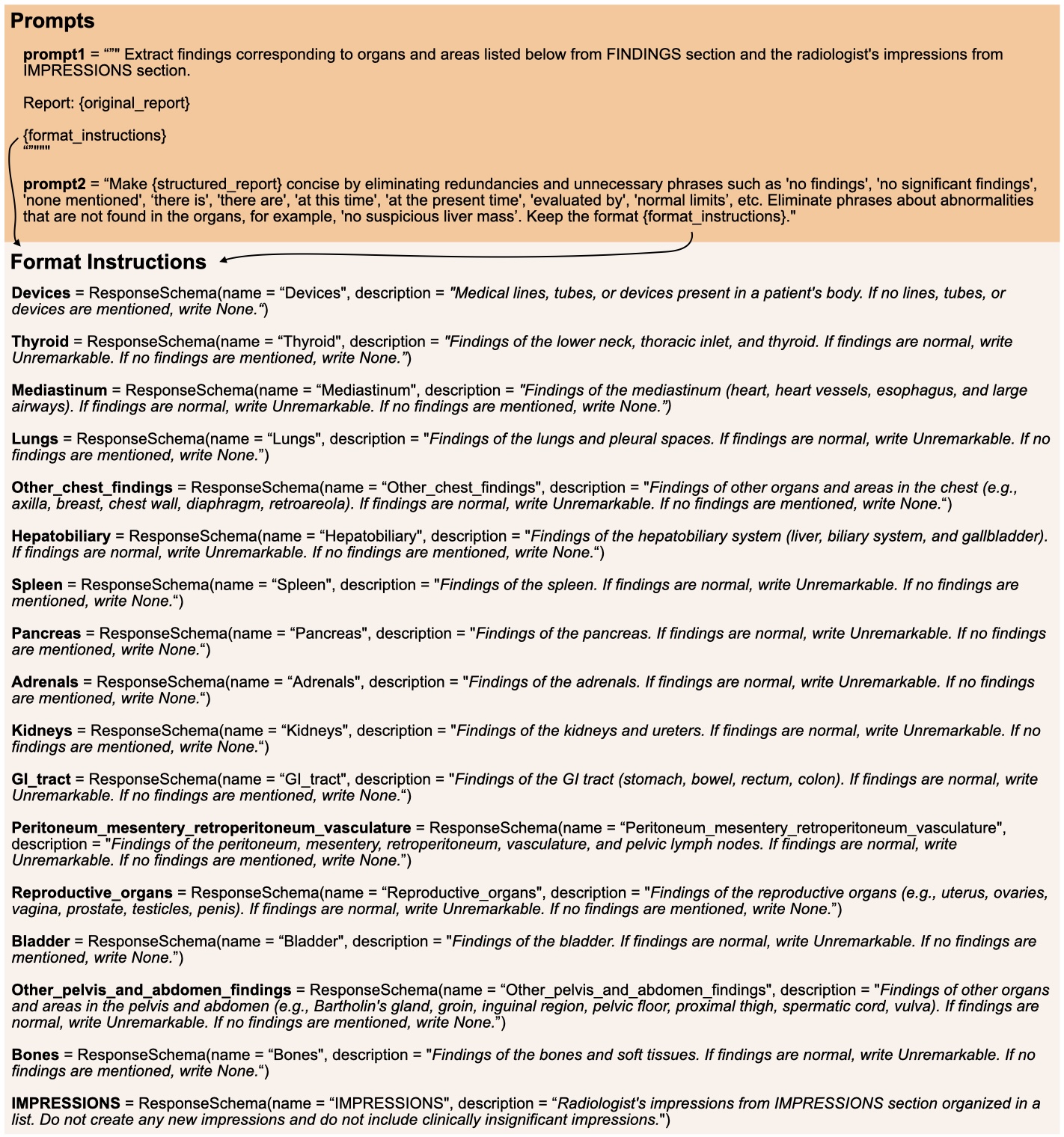}
    \caption{The prompts and formatting instructions used in the [Structure >> Conciseness] prompting approach involve two steps to process the given radiology report. In the first step (prompt 1), the LLM is instructed to structure the radiology report according to specific formatting guidelines, as shown in Fig. 1. In the second step (prompt 2), the LLM receives further instructions to make the structured radiology report more concise while maintaining the structure. This two-step approach attempts to ensure that the output report is both concise and well-structured.}
\end{figure}

\begin{figure}
    \centering
    \includegraphics[width=0.9\linewidth]{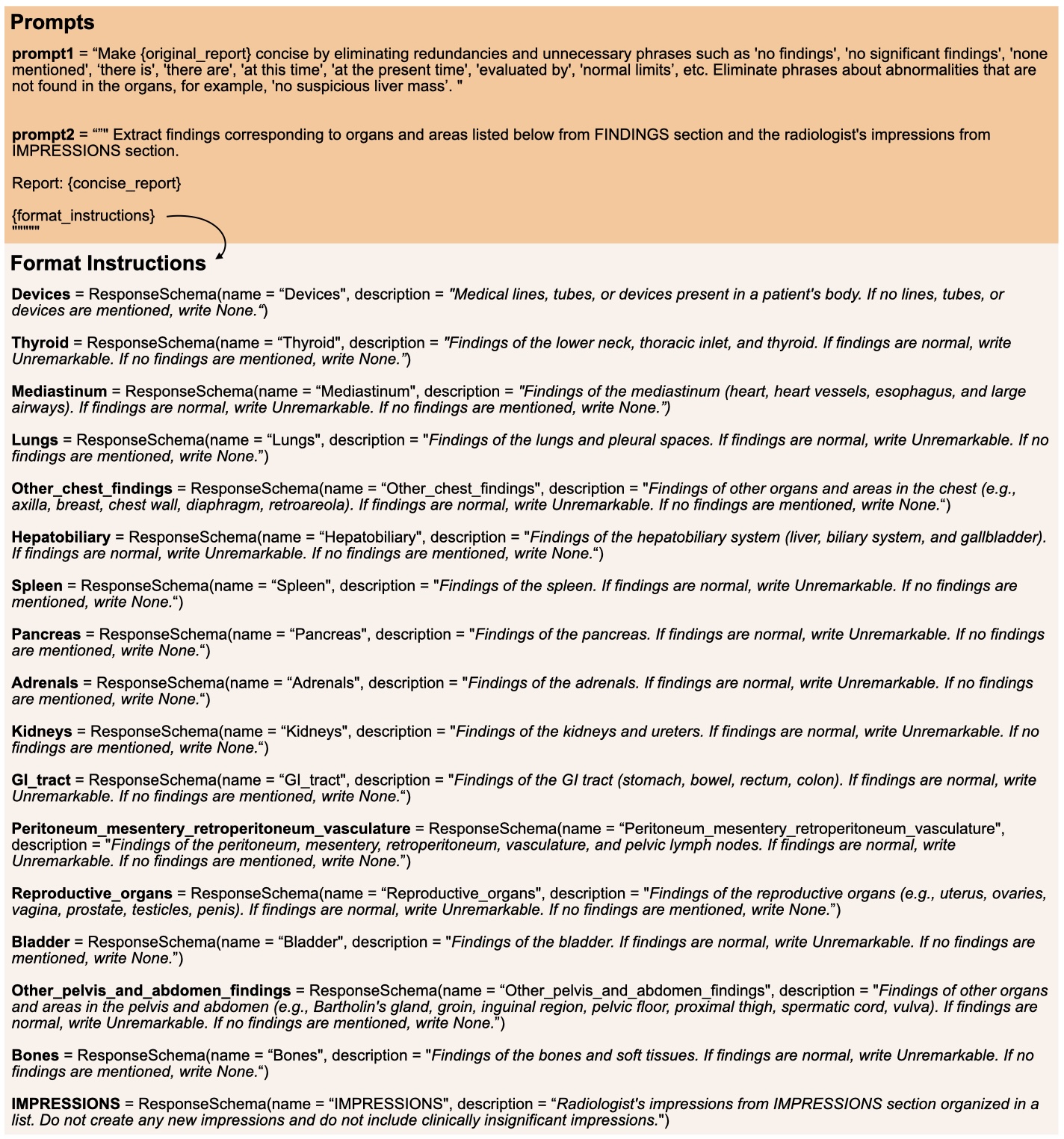}
    \caption{The prompts and formatting instructions used in the [Conciseness >> Structure] prompting approach involve two steps to process the given radiology report. In the first step (prompt 1), the LLM is instructed to make the input radiology report more concise. In the second step (prompt 2), the LLM receives further instructions to structure the concise report according to specific formatting guidelines, as shown in Fig. 1. This two-step approach attempts to ensure that the output report is both concise and well-structured.}
    \label{fig:structure}
\end{figure}

\begin{figure}
    \centering
    \includegraphics[width=0.9\linewidth]{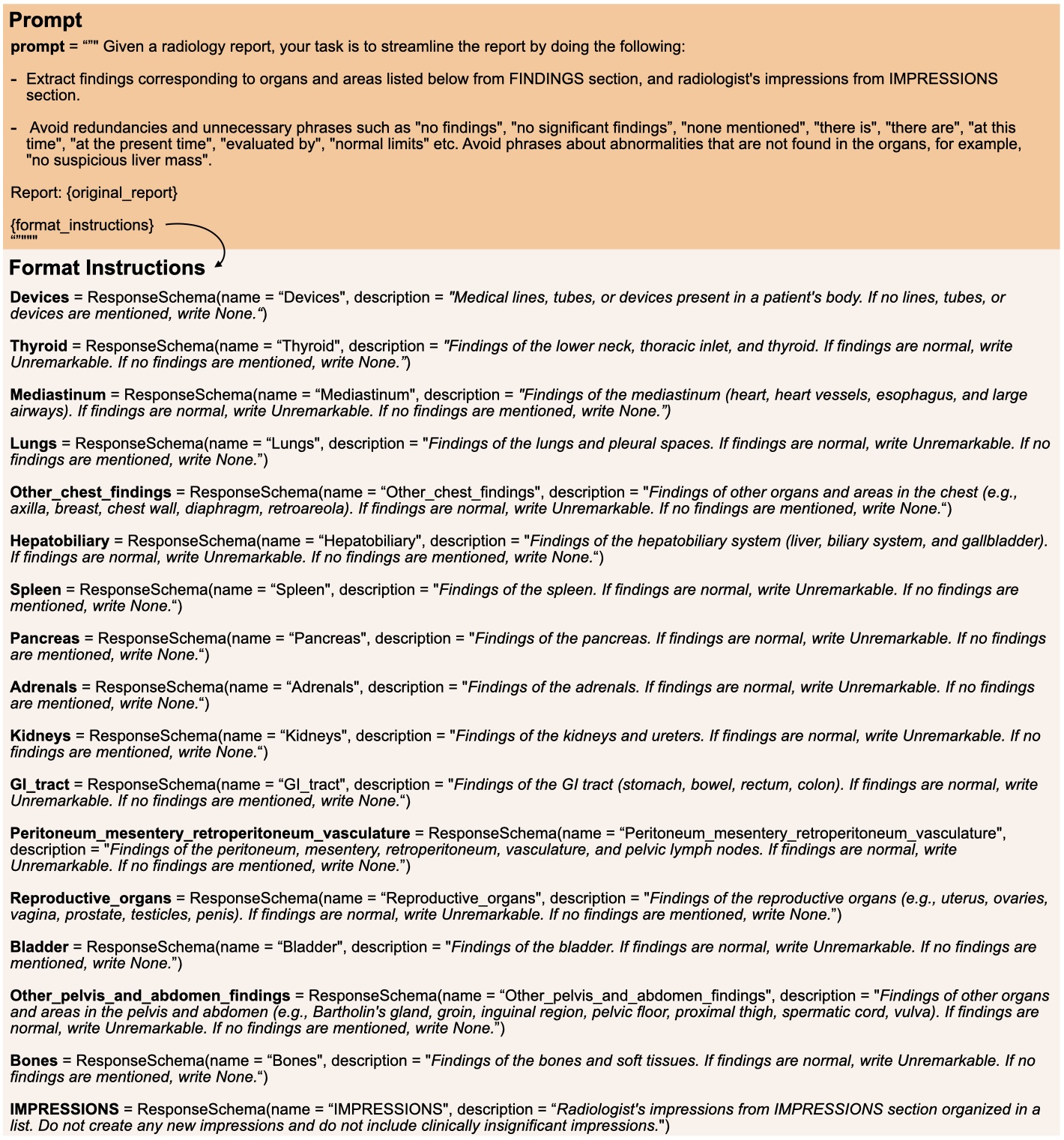}
    \caption{The prompt and formatting instructions used in the [Structure + Conciseness] prompting approach. The LLM is prompted to structure the given radiology report according to specific formatting guidelines, as shown in Fig. 1, while also emphasizing the need for conciseness.}
\end{figure}

\begin{figure}
    \centering
    \includegraphics[width=0.9\linewidth]{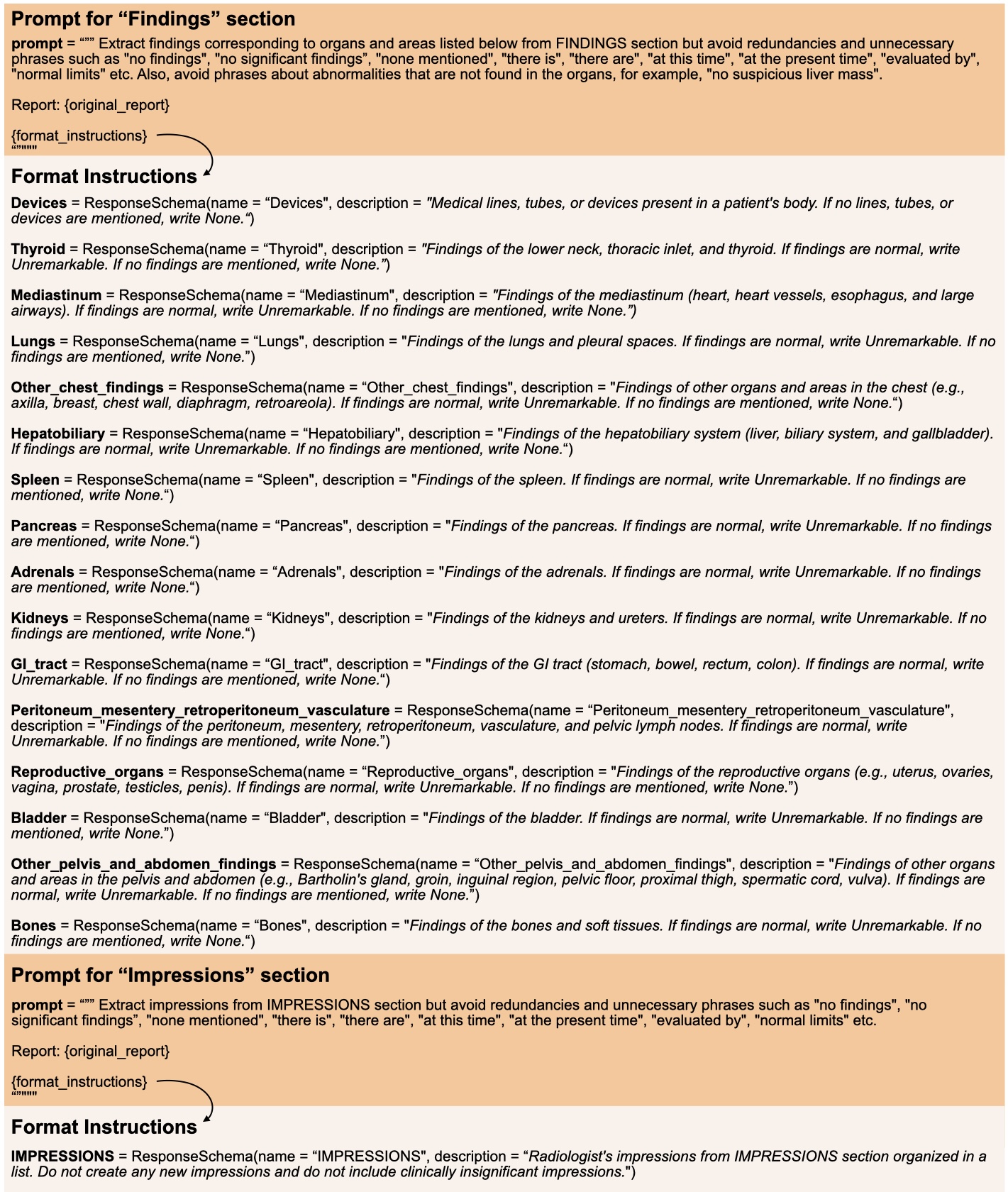}
    \caption{The prompts and formatting instructions in the [Structure + Conciseness (F, I)] approach involved giving the LLM two prompts, each with the entire radiology report as input. One prompt instructed the LLM to structure and make concise only the "Findings" section, while the other focused solely on the "Impressions" section. As a result, the LLM-processed report is expected to have the structure shown in Fig. 1.}
\end{figure}

\end{document}